\documentclass[letterpaper, 10 pt, conference]{ieeeconf}  

\IEEEoverridecommandlockouts                              

\usepackage{lipsum} 
\usepackage{amssymb,graphicx,amsmath,color}
\usepackage{bm}
\usepackage{subcaption}
\usepackage{graphicx}
\usepackage{epstopdf}
\usepackage{placeins}
\usepackage{gensymb}
\usepackage[hidelinks]{hyperref}
\usepackage[percent]{overpic}
\usepackage{tabularx}
\usepackage{threeparttable}
\usepackage{courier} 
\usepackage{float}
\usepackage{algorithm}
\usepackage{algpseudocode}

\usepackage[noadjust]{cite}

\title{\LARGE \bf
Principal Components of Touch
}
 
\author{Kirsty Aquilina$^{1}$,  David A. W. Barton$^{2}$ and Nathan F. Lepora$^{2}$
\thanks{The work of KA was supported by the EPSRC Centre for Doctoral Training in Future Autonomous and Robotic Systems (FARSCOPE) at the Bristol Robotics Laboratory. The work of NL was supported in part by a Leadership Award from the Leverhulme Trust on `A biomimetic forebrain for robot touch' (RL-2016-39).}
\thanks{Data used in this paper is available at http://doi.org/ckqw}
\thanks{$^{1}$KA is with Bristol Robotics Laboratory, University of Bristol, UK.}%
\thanks{$^{2}$DB and NL are with the Department of Engineering Mathematics and Bristol Robotics Laboratory, University of Bristol, Bristol, UK.\newline
    Email: \{ka14187, david.barton, n.lepora\}@bristol.ac.uk%
    }%
}

\begin{document}

\maketitle
\thispagestyle{empty}
\pagestyle{empty}


\begin{abstract}
Our human sense of touch enables us to manipulate our surroundings; therefore, complex robotic manipulation will require artificial tactile sensing. Typically tactile sensor arrays are used in robotics, implying that a straightforward way of interpreting multidimensional data is required. In this paper we present a simple visualisation approach based on applying principal component analysis (PCA) to systematically collected sets of tactile data. We apply the visualisation approach to 4 different types of tactile sensor, encompassing fingertips and vibrissal arrays. The results show that PCA can reveal structure and regularities in the tactile data, which also permits the use of simple classifiers such as $k$-NN to achieve good inference. Additionally, the Euclidean distance in principal component space gives a measure of sensitivity, which can aid visualisation and also be used to find regions in the tactile input space where the sensor is able to perceive with higher accuracy. We expect that these observations will generalise, and thus offer the potential for novel control methods based on touch.

\end{abstract}


\section{INTRODUCTION}

Our human sense of touch enables our capabilities to manipulate the world around us~\cite{johansson2009}, taking inspiration from humans complex robotic manipulation should also be equipped with artificial tactile sensing. Typically, tactile sensors comprise arrays of sensing elements called taxels (see {\em e.g.}~\cite{kappassov2015} for a review). Therefore there is a need for straightforward interpretation of multi-dimensional time series data for tactile sensing applications, which is a difficult problem because the transduction of physical contact into multi-dimensional time series can be highly complex.

 \begin{figure}[t!]
	\begin{center}
		\begin{overpic}[width=0.35\textwidth]{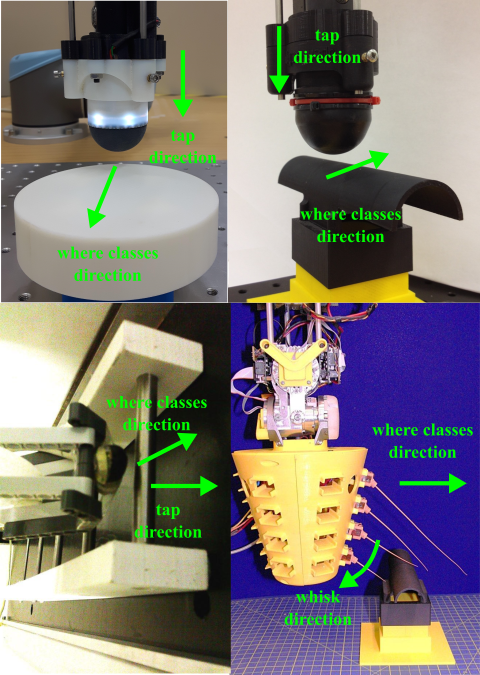}
			\put(2,95){\color{green}\textbf{(a)}}
			\put(34,95){\color{green}\textbf{(b)}}
			\put(2,48){\color{green}\textbf{(c)}}
			\put(34,48){\color{green}\textbf{(d)}}
		\end{overpic}
	\end{center}
	\caption{Tactile experimental platforms. (a) Optical tactile sensor (TacTip~v2) mounted on UR5 6-DOF robot arm. (b) TacTip~v1 mounted on ABB 6-DOF robot arm. (c) Capacitive tactile sensor (iCub fingertip) on a Yamaha 2-DOF Cartesian Robot. (d) BIOTACT whisker array on an Elumotion 7-DOF robot arm. (b)-(d) were considered previously in ref.~\cite{Lepora2016}.}
	\label{setups}
	\vspace{0.55em}
\end{figure}
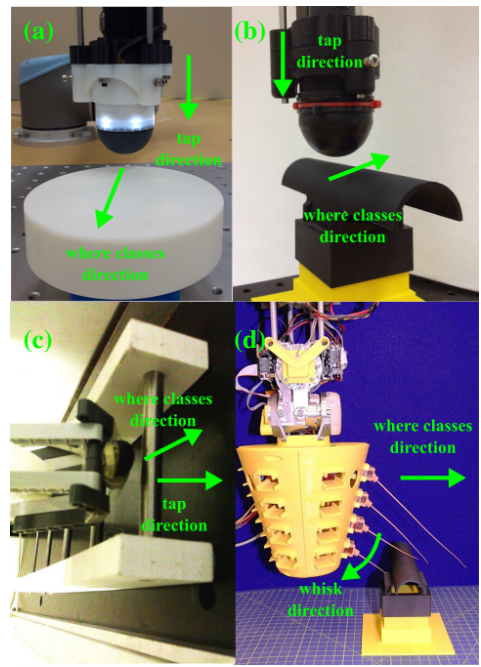

In this paper we present a simple visualisation approach based on applying principal component analysis (PCA) to systematically collected sets of tactile data. We find a surprising amount of regularity in the manifolds formed by this linear dimensional reduction technique, from which we are able to directly visualise quantities of physical relevance, such as the curvature of the stimulus or the contact location. 

To demonstrate generality, this method is applied to four different tactile sensor arrays (Fig. \ref{setups}), encompassing two types of optical tactile sensor (known as TacTip~v2 and TacTip~v1~\cite{chorley2009,ward2018tactip}), a capacitive tactile sensor (the iCub fingertip)~\cite{schmitz2011} and a biomimetic whisker array~\cite{sullivan2012}. Each sensor has a discrete number of taxels with overlapping receptive fields, making these sensory dimensions highly correlated. This correlation implies that dimensionality reduction techniques can uncover the underlying regularity within an entire array of activated taxels.

An additional benefit of this approach is that a sensitivity value for each dimensionally-reduced data element can be computed to measure the distinguishability between different stimuli and contact locations in that region of the data set. This has direct relevance to active tactile perception~\cite{prescott2011,lepora2016scholarpedia} because it can reveal how best to relocate the tactile sensor to optimize the perceptual acuity~\cite{Lepora2015}. 

\section{BACKGROUND}

This paper investigates how information can be obtained from tactile data using dimensionality reduction, focussing on principal component analysis (PCA). The most common application of PCA to robot touch has been for a low-dimensional feature extraction preprocessing step before classification~\cite{corradi2015,Eguiluz2016,Goger2009}; such approaches have been applied to Zernike moments of the data~\cite{corradi2015}, to the Fast Fourier Transform of the data~\cite{Eguiluz2016,Goger2009} and on the full tactile images (matrix of pressure values at each taxel)~\cite{Goger2009}. Another application has been for estimating object pose from touch, by applying PCA to tactile data that is then matched with a point cloud of an object of interest~\cite{Bimbo2016}.  

Our interest also encompasses the visualization of tactile data. A common visualisation approach for tactile perception is to show a snapshot of a tactile image when contacting a specific stimulus~\cite{Bimbo2016,Goger2009,li2013}. Another approach~\cite{cannata2010} creates a mesh called a 2D somatosensory map to represent the contact location on a tactile skin. A self-organising map has also been used to visualise tactile data for various objects, by depicting the clusters formed from different stimuli~\cite{Natale06}. The work which is most similar to the present study uses t-SNE to visualise the tactile data for a force and curvature experiment~\cite{Karl2016ml}; however, there the visualisation is used to explain the results obtained after performing inference, whereas here we consider the opposite approach of using the visualization to aid inference.

\section{METHODS}
\begin{figure*}
		\vspace{0.35 em}
		\centering
		\includegraphics[width = 0.95\textwidth]{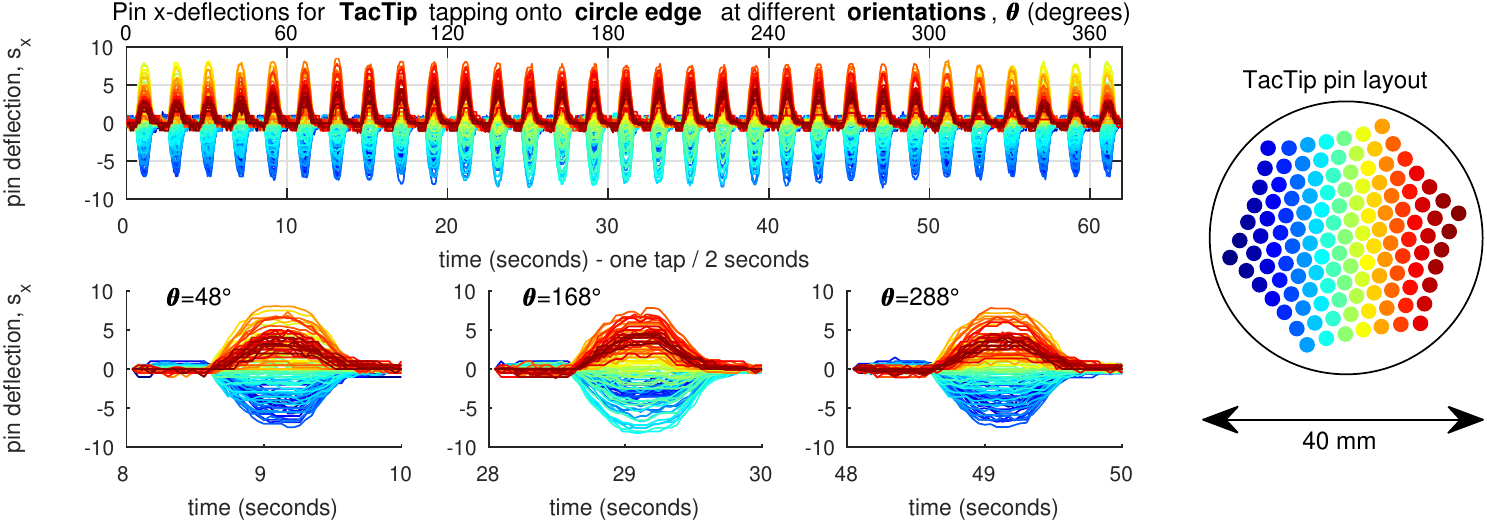}
		\caption{Top: The sensor measurements obtained when the TacTip~v2 taps over the edge of the circular stimulus at all the different orientations. Bottom: Sensor measurements obtained during taps on the edge at 3 different orientations (48\degree{}, 168\degree{} and 288\degree{}). The colours of each timeseries corresponds to the taxel colour shown on the TacTip pin Layout (right) and the magnitude of each timeseries is the x-deflection of each pin, a similar plot is obtained when considering the y-deflections. Refer to ref. \cite{Lepora2016} for sensor measurements plots of TacTip~v1, iCub fingertip and BIOTACT whisker array.}
		\label{fig:rawData}
\end{figure*}
\subsection{Experimental Setups}

Three of the tactile experiments considered in this paper (Fig.~\ref{setups}b-d) are from a previous study of biomimetic active touch with tactile whiskers and fingertips~\cite{Lepora2016}, with one new dataset gathered specifically for the present study (Fig.~\ref{setups}a). The data gathered from each experiment has distinct training and test sets for use in validation.

All collected data is labelled by a `where' $x_l$ class for stimulus location and a `what' $w_i$ class for stimulus identity. The data is collected in a systematic grid where for each `what' $w_i$ class the same number of `where' classes are collected. Typically, the `where' classes refer to lateral positions, as shown in Fig.~\ref{setups}, and the `what' classes are distinct objects such as cylinder diameter or edge orientation. 

The tactile experiments and collected data are as follows:

\subsubsection{TacTip~v2 (Fig.~\ref{setups}a)} \label{sec:expTactip2}
The TacTip~v2 is an improved version of the original TacTip~v1 optical tactile sensor~\cite{chorley2009}. The sensor is fully 3D-printed with 127 pins arranged in a hexagonal array inside a 40\,mm diameter compliant dome that is filled with a gel. The motion of these pins is tracked with a USB camera (Microsoft Lifecam). The projected hexgaonal pattern results in evenly distributed pins in the camera image. TacTip~v2 was mounted on a UR5, a 6-DOF robot arm, for the study in this paper.

The experiment involved tapping the tactile sensor against the edge of a circular object, while systematically varying the orientation and radial displacement of the sensor relative to the edge. The orientation `what' classes range from 0\degree{}--360\degree{} in 12\degree{} increments and the radial positions `where' classes from -12\,mm--5\,mm in 0.5\,mm increments. The experiment started in free space and ended on the object surface, with the origin at 0\,mm denoting the object edge. The sensor measurements obtained during taps over the stimulus edge whilst varying the sensor orientation are depicted in Fig.~\ref{fig:rawData}.

\subsubsection{TacTip~v1 (Fig.~\ref{setups}b)} \label{sec:expTactip1}
The original TacTip sensor~\cite{chorley2009} (denoted as v1) differs from its more recent version by having a moulded rubber like skin with 532 pins arranged in a geodesic pattern. The data gathered with this sensor used only a subset of 38 pins, as described in the original study~\cite{Lepora2016}. For data collection, the TacTip~v1 was mounted on a IRB120 ABB 6-DOF robot arm~\cite{Lepora2016}. 

The data collection involved tapping the sensor against 6 distinct cylinders, comprising the `what' identity classes labelled by the cylinder diameter (curvature diameter) 30\,mm--80\,mm in 10\,mm increments; these contacts were taken over a 40\,mm range of lateral displacement `where' classes in 0.04\,mm increments. The sensor started with no contact, moved onto the cylinder, and finished with no contact. 

\subsubsection{iCub Fingertip (Capacitive Tactile Fingertip) (Fig.~\ref{setups}c)}
This capacitive tactile sensor was fabricated for the iCub humanoid~\cite{schmitz2011}. Its shape is similar to a human fingertip, with 12 taxels embedded within silicon foam and a soft rubber outer surface. For data collection, the iCub fingertip  was mounted on a 2-axis PXYx Cartesian robot (Yamaha Robotics).

The sensor tapped against 5 cylinders, comprising the `what' classes with diameters ranging from 4\,mm--12\,mm in 2\,mm increments; these contacts were taken over a 30\,mm range of lateral displacements `where' classes in 0.01\,mm increments. The sensor started by tapping its rigid insensitive base, then contacted the object, and finished with no contact.

\subsubsection{Tactile Whiskers (BIOTACT Vibrissae) (Fig.~\ref{setups}d)} \label{sec:expWhisk}
The BIOTACT whisker array is inspired from the rodent snout~\cite{sullivan2012}, and jointly developed by Bristol Robotics Lab and the University of Sheffield. The whiskers were manufactured in a tapered shape mimicking the shape of rodent whiskers; the sensor measures their deflection using Hall-effect sensors at the whisker base, with here only 4 of the whiskers in the array being used~\cite{Lepora2016,Lepora2012icra}. This sensor array was mounted on an El-arm Elumotion 7 DOF robot arm. 

The data collection again involved an experiment in which the tactile whiskers are tapped (whisked) against cylindrical stimuli. The same stimuli as the TacTip~v1 experiment were used, with 6 `what'  cylinder diameter (curvature diameter) classes; meanwhile the lateral displacement `where' classes ranged over 100\,mm in increments of 0.25\,mm.

\subsection{Algorithms}

\subsubsection{Data Representation}

The data used in this study is recorded in discrete segments, specifically during taps onto, then off, the stimulus of interest. Each segment of data is a multi-dimensional time series
\begin{equation}
\label{multi dim data}
z =\{s_k(j)\ :\ 1\leq j\leq N_{\rm samples},\ 1 \leq k \leq N_{\rm dims}\},
\end{equation}
with $j$ the time sample index for a segment and $k$ the dimension index; here $N_{\rm samples}$ is the number of time samples per segment and $N_{\rm dims}$ is the total number of sensor dimensions.

Each experiment has a total of $N =N_{\rm id}\times N_{\rm loc} $ perceptual classes where $N_{\rm id}$ is the number of `what' identity classes and $N_{\rm loc}$ is the number of `where' location classes, with each data segment $z$ having a $(x_l,w_i)$ label. A summary of the experiment data details is provided in \autoref{tbl:Experiment data}. 

\begin{table}[]
	\centering
	\caption{Experiment data details}
	\label{tbl:Experiment data}
	\begin{threeparttable}
		\setlength{\tabcolsep}{0.4em}
		\begin{tabular}{ccccc}
		\hline
		& TacTip v2& TacTip v1 & iCub & Whiskers \\ \hline
		taxels & 127&      38     & 12   & 4 \\
		$N_{\rm dims}$  & 254\tnote{*}&  76\tnote{*}  & 12   & 8\tnote{*} \\
		$N_{\rm samples}$  &46&47&51   & 1000 \\
		\begin{tabular}[c]{@{}c@{}}what class\\ $N_{\rm id}$\\ range\\ increment\end{tabular}   &  \begin{tabular}[c]{@{}c@{}}orientation \\ 31\\0\degree{}--360\degree{} \\ 12\degree{}\end{tabular} & \begin{tabular}[c]{@{}c@{}}cylinders \\ 6\\30--80\,mm \\ 10\,mm\end{tabular}&\begin{tabular}[c]{@{}c@{}}cylinders \\ 5\\4--12\,mm \\ 2\,mm\end{tabular}&\begin{tabular}[c]{@{}c@{}}cylinders \\ 6\\30--80\,mm \\ 10\,mm\end{tabular} \\
		\begin{tabular}[c]{@{}c@{}} where class \vspace{1em} \\$N_{\rm loc}$\\ range\\ increment\end{tabular} & \begin{tabular}[c]{@{}c@{}}radial \\ displacement \\ 35\\-12--5\,mm \\ 0.5\,mm\end{tabular} &      \begin{tabular}[c]{@{}c@{}}lateral \\ displacement \\ 1000\\0--40\,mm \\ 0.04\,mm\end{tabular}& \begin{tabular}[c]{@{}c@{}}lateral \\ displacement \\ 3000\\0--30\,mm \\ 0.01\,mm\end{tabular}& \begin{tabular}[c]{@{}c@{}}lateral \\ displacement \\ 400\\0--100\,mm \\ 0.25\,mm\end{tabular}\\ \hline
		\end{tabular}
		\begin{tablenotes}
			\item[*] 2 dimensions per taxel
		\end{tablenotes}
	 \end{threeparttable}
\end{table}

\subsubsection{Principal Component Analysis}

A linear dimensionality reduction technique, namely PCA, is used to obtain the overall information content from the multi-dimensional time series $z$ for each tap. This transformation of the input data to a lower dimensional manifold eases visualisation and effectively filters input noise \cite{rosipal2001kernel}. The data is first preprocessed by removing taps with abnormally large or small numbers of time samples and centering each sensor dimension time series in each tap. PCA is used on the preprocessed data to find the directions in which the data has maximum variance and project the data onto those orthogonal vectors ({\em e.g.}~\cite{Jolliffe2002}).
 
In practice, the data is organised as a matrix $[s_{n,j}]_k$ of dimensions $N\times N_{\rm samples}$ for each sensor dimension $k$ (with $n$ the row-index from reordering the classes $(l,i)$, where $1 \leq n \leq N$). PCA is performed on each matrix to obtain the eigenvectors $\bm{y}_{k,j}$ along the $j$-th principal direction, an $N_{\rm samples}$-dimensional eigenvector that projects the data into its principal components (PCs). These PCs are ordered in decreasing eigenvalue (variance) $\lambda_{k,j}$ keeping up to the last PC with
\begin{equation}
\label{eq:scree}
\frac{\lambda_{k,j}-\lambda_{k,(j+1)}}{\sum_{j=1}^{N_{\rm samples}}\lambda_{k,j}}> \gamma
\end{equation}
where $\gamma$ is a threshold here set to 0.05. This results in a time-compressed data representation of dimension $N\times N_{\rm total}$ where $N_{\rm total}$ is the total number of dimensions satisfying~\eqref{eq:scree}.
PCA is then performed again on the time-compressed data with the same criterion~\eqref{eq:scree} ($\gamma$ set to 0.005), applied to determine the number of PC dimensions $ N_{\rm reduced}$ to keep. This resulted in an $N_{\rm reduced}$-dimensional vector $\bm{p}$ for each data segment $z$. The PCs $\bm{p}$ describe the spatial variance of the taxels, with the compound PCA projection transforming the collection of input data $z(x_l,w_i)$ into a collection of PCs, which lie within a surface ${\cal M}_P$.

\subsubsection{Nearest Neighbour Classifier}
\label{sec:methodsKnn}
A $k$-nearest neighbour (with $k=1$) approach can then be used to construct a very basic classifier. The classifier uses the projected test data $z$ into the surface $\cal{M}_P$ as an input. The classifier searches for the nearest neighbour $(x_l,w_i)$ in its training set using the Euclidean distance as a measure of similarity. Here two classifiers are constructed: one to infer the `what' identity class $w_i$ and another to infer the `where' location class $x_l$, by assigning the appropriate label to each observation in the training data. The $k$-NN classifier is implemented with an exhaustive search algorithm and using uniform prior probabilities for each class (implemented with `fitcknn' from the MATLAB statistics and machine learning toolbox).

\subsubsection{Probabilistic Classifier}
\label{sec:probClassifier}

A simpler version of probabilistic classifier previously applied to three of the present datasets~\cite{Lepora2016} is also used for comparison, corresponding to a maximum marginal mean likelihood classifier over a histogram likelihood model. The probabilistic classifier builds a histogram likelihood model for each class in the training set, which results in the probability distribution $P_k(s_k|x_l,w_i)$. 
In this paper the dimensionally-reduced data are presented to the classifier which results in
\begin{equation}
\label{eq:likelihoodPCA}
\log P(z|x_l,w_i) =  \sum_{r=1}^{N_{\rm reduced}}\frac{\log P(p_r|x_l,w_i)}{N_{\rm reduced}}.
\end{equation}
where $r$ is the principal component index $1\leq r\leq N_{\rm reduced}$ for each PC vector $\bm{p}$. Then the most likely `what' class
\begin{eqnarray}
\label{eq:MAP}
w_{\rm dec} = \operatorname*{arg\,max}_{w_i}  \sum_{l=1}^{N_{\rm loc}} P(z|x_l,w_i)
\end{eqnarray}
corresponds to the maximal marginalised probabilities.

\subsubsection{Sensitivity Computation}
\label{sec:sensitivityMethod}
Using the projected surface $\cal{M}_P$ from applying the PCA to the training data, one can estimate a measure of sensitivity for each element of the training dataset. Sensitivity here denotes the ability of the sensor to distinguish between different projections $\bm{p}$ of the data. Specifically, this can be described as the change in principal component value $\Delta\bm{p}$ produced by a unit change in the measured quantity, here corresponding to the `what' identity class $w_i$ \cite[p.\,17]{Kalantar-zadeh2013}:
\begin{equation}
\label{eq:sens}
\text{Sensitivity}= \frac{||\Delta\bm{p}||}{|\Delta w_i|}
\end{equation}
The larger the sensitivity, the easier it is for the sensor to distinguish between different contacts. 

Here we compute the sensitivity for each data element $z$ using \autoref{alg:overall} (with overlap parameter $b=10$). \autoref{alg:distance} is used within \autoref{alg:overall} with $\theta_{\text{threshold}} = \pi/18$ for the distance computation required for the sensitivity of each training data point. The sensitivity is then used to find a suitable `where' location class  (referred to as fixation point) that results in the best perception for the given setup. The chosen sensitivity measure was to sort in ascending order the minimum and maximum filtered sensitivity value at each `where' location class $x_l$ (therefore finding the maximum and minimum of the sensitivities for all the identity classes at that location) and choosing the location class which has the largest overall sum of the two rankings.
This measure was chosen using a 10 fold cross validation (this used only data present in the training set).

\begin{algorithm}
	\caption{Find most distinguishable location}\label{alg:overall}
  	\begin{algorithmic}[1]
		\Require all training vectors $\bm{p}$ , labels 
		\Ensure fixation point
		\State Sort data according to $p_1$
		\State Divide data in $b$ overlapping sections
		\State computeDistance() for each point
		\State sensitivity = Distance/$\Delta$Label
		\State fixation point = chosen sensitivity measure
	\end{algorithmic}
\end{algorithm}
\begin{algorithm}
	\caption{computeDistance()}\label{alg:distance}
	\begin{algorithmic}[1]
		\Require Sorted Batches of training data $\bm{p}$, labels 
		\Ensure distance
		\State Find the vectors between all points with different labels
		\State Find the shortest vector $v_{\text{close}}[1]$
		\State  $N_\text{neighbours} = 1$ 
		\While {$N_\text{neighbours} < 5$}
		\State  compute  $(\hat{v}_{\text{close}}.\hat{v}) >\cos(\theta_{\text{threshold}} ) ~ \forall~  v$
		\State find next shortest vector $v$ that satisfies condition 5
		\State remove all vectors that do not satisfy condition 5
		\State $v_{\text{close}}[N_\text{neighbours}]=v$
		\State $N_\text{neighbours} = N_\text{neighbours}+1$
		\EndWhile
		\State distance = min(median($v_{\text{close}}$) , previous value from neighbouring section)
	\end{algorithmic}
\end{algorithm}

\section{RESULTS}
\subsection{Visualisation of Principal Components}
\label{sec:resultVis}
\begin{figure*}[t!]
	\vspace{0.7em}
	\centering
	
	\begin{subfigure}{0.22\textwidth}	
		\centering
		\begin{overpic}[width=\textwidth]{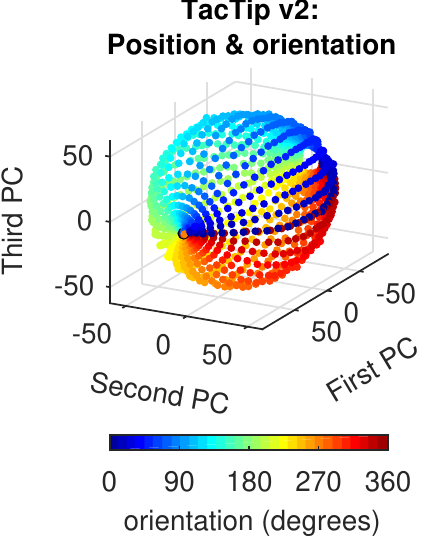}
			\put(5,97){\textbf{(a)}}
		\end{overpic}
		\label{PCA-tactipv2-what}
	\end{subfigure}	\hfill
	\begin{subfigure}{0.22\textwidth}
		\centering
		\begin{overpic}[width=\textwidth]{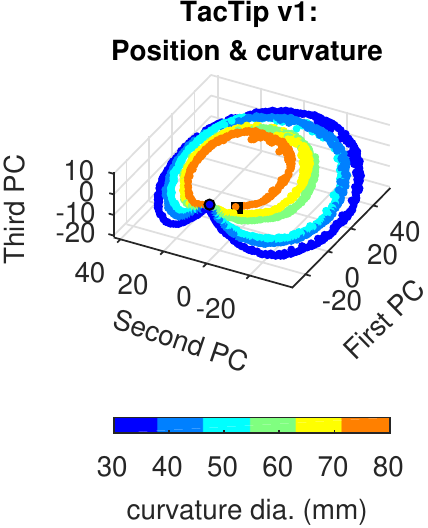}
			\put(5,97){\textbf{(b)}}
		\end{overpic}
		\label{PCA-tactipv1-what}
	\end{subfigure}	\hfill
	\begin{subfigure}{0.22\textwidth}
		\centering
		\begin{overpic}[width=\textwidth]{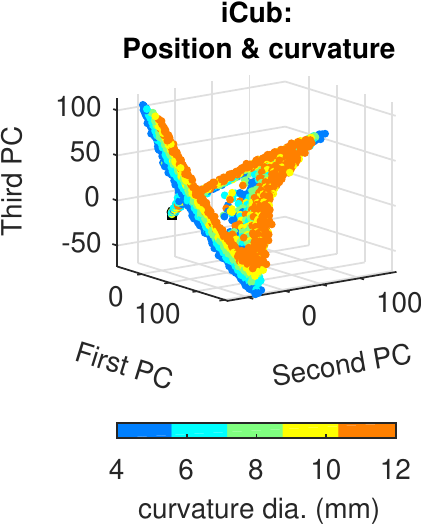}
			\put(5,97){\textbf{(c)}}
		\end{overpic}
		\label{PCA-icub-what}
	\end{subfigure}	\hfill
	\begin{subfigure}{0.22\textwidth}
		\centering
		\begin{overpic}[width=\textwidth]{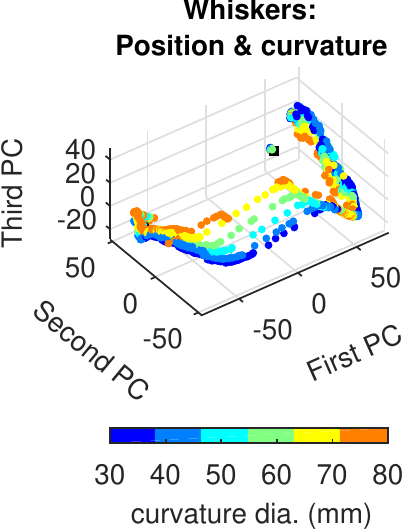}
			\put(5,97){\textbf{(d)}}
		\end{overpic}
		\label{PCA-biotact-what}
	\end{subfigure}

	\begin{subfigure}{0.22\textwidth}
		\centering
		\begin{overpic}[width=\textwidth]{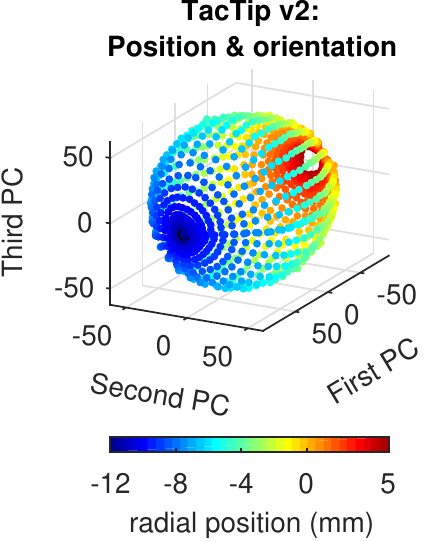}
			\put(5,97){\textbf{(e)}}
		\end{overpic}
		\label{PCA-tactipv2-where}
	\end{subfigure}	\hfill
	\begin{subfigure}{0.22\textwidth}
		\centering
		\begin{overpic}[width=\textwidth]{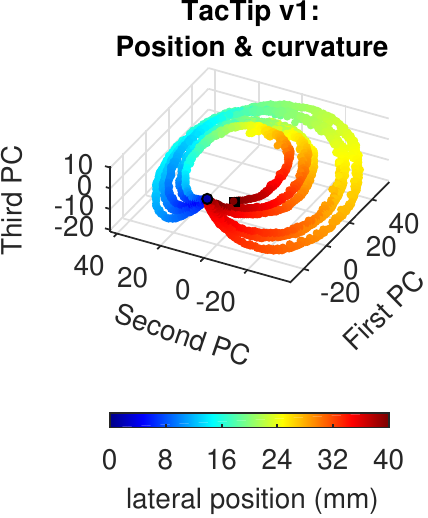}
			\put(5,97){\textbf{(f)}}
		\end{overpic}
		\label{PCA-tactipv1-where}
	\end{subfigure}	\hfill
	\begin{subfigure}{0.22\textwidth}
		\centering
		\begin{overpic}[width=\textwidth]{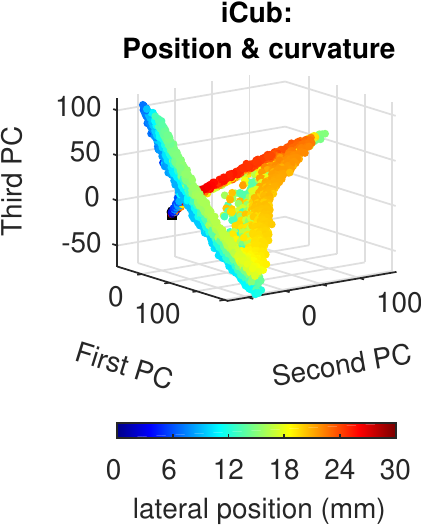}
			\put(5,97){\textbf{(g)}}
		\end{overpic}
		\label{PCA-iCub-where}
	\end{subfigure}	\hfill
	\begin{subfigure}{0.22\textwidth}
		\centering
		\begin{overpic}[width=\textwidth]{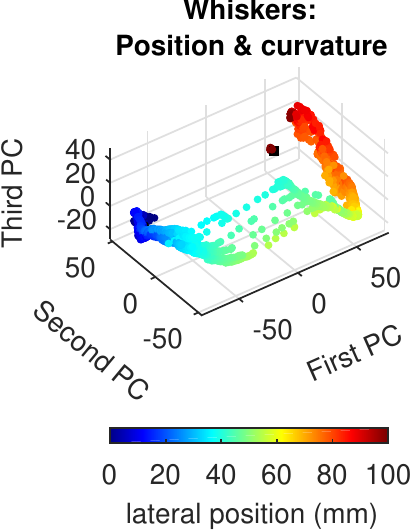}
			\put(5,97){\textbf{(h)}}
		\end{overpic}
		\label{PCA-biotact-where}
	\end{subfigure}
	\vspace{-1em}
	\caption{Tactile principal components. The first three dimensions of the low dimensional manifold ${\cal M}_P$ are portrayed here. The colour of each PC data point changes according to its class label. (a,e) TacTip~v2 orientation from smallest angle (0$^\circ$, blue) to largest angle (360$^\circ$, red) and radial position from no contact (-12\,mm, blue) to full contact (5\,mm, red). (b,f) TacTip~v1, (c,g) iCub fingertip and (d,h) BIOTACT whiskers experiments over cylinder curvature from smallest (blue) to largest (orange) diameter and lateral position from left (blue) to right (red) extremity. The gradual changes in colour show the structure and smoothness of the low dimensional principal component manifold.}
	\label{PCA}
\end{figure*}

Our main aim is to visualise the tactile data collected over an entire experiment in which the `where' contact location $x_l$  and `what' stimulus identity $w_i$ vary systematically over discrete classes. Each labelled data $z(x_i,w_l)$ (comprising a single tap) is projected into 3-dimensions corresponding to the leading principal components (Fig.~\ref{PCA}), with each point coloured according to its `what' $w_i$ or `where' $x_l$ class.

Considering first the TacTip~v2 orientation and radial displacement experiment (Fig.~\ref{setups}a; Sec.~\ref{sec:expTactip2}), we can observe a surprising amount of regularity and structure in the manifold created by the PCA projection over the entire where-what dataset (Fig.~\ref{PCA}a,e). The sensor orientation directly relates to the PCA vector angle (Fig.~\ref{PCA}a) in the PC$_2$-PC$_3$ plane, as evident from the coloured `what' orientation class (Fig.~\ref{PCA}a). Meanwhile, the radial position of the sensor relates to the translations of the PCA vector in the PC$_1$-PC$_3$ plane (Fig.~\ref{PCA}e). Note also how the PCA vectors concentrate around a single point on the graph when the sensor is in free space and the data becomes indiscriminable over the orientation.  Even so, it is interesting to note that the PC vectors with largest radial magnitude are obtained just before the sensor contacts the edge (Fig.~\ref{PCA}e; 11\textsuperscript{th}-20\textsuperscript{th} classes corresponding to -7\,mm to -2\,mm). We interpret that when a large part of the sensor surface contacts the stimulus, the mobility of the pins is reduced, resulting in more difficult orientation perception. The structure in the data shown through the visualisation is therefore a good way to understand the interaction between the sensor and the stimulus.

An aspect of the PC visualization is that the magnitude of each PC vector directly relates to the magnitude of the taxel displacements during a contact. Close to free space  when most of the sensor surface is off the stimulus, the taxel displacements and hence PC vectors are smaller than directly on the edge of the stimulus, making perception difficult. Additionally, when  the sensor is further onto the object the difference in PC vectors decreases as the sensor location becomes indistinguishable (Fig.~\ref{PCA}e).

Consider next the three other tactile sensors (TacTip~v1, iCub fingertip and BIOTACT whisker array from Figs. 1b-d) performing similar experiments of tapping against a range of cylinder diameters $w_i$ over a span of locations $x_l$, comprising the `what' and `where' identity and location classes ( Sec.~\ref{sec:expTactip1}--Sec.~\ref{sec:expWhisk}). In all cases, the projected PCs vary systematically with cylinder diameter (Figs. \ref{PCA}b-d; orange to blue denotes largest to smallest). The PCA vectors have a larger magnitude for the smallest cylinder, showing that it is more distinguishable over location than the larger cylinders (Figs.~\ref{PCA}b,f), as expected from the relative curvature. Once again, the strength of contact directly relates to the magnitude of the PCA vector, which can be clearly seen in the visualisation vectors (Figs.~\ref{PCA}f-h) varying according to the `where' lateral  location class. Again the data converges when the sensor reaches free space, relating to the ambiguity over the `what' class label $w_i$.  

The iCub fingertip (Fig.~1c, Figs.~\ref{PCA}c,g) is unique in this study in having each sensor dimension a single taxel, so the sensor output is in fact a tactile image. The visualisation (Fig.~\ref{PCA}c) reveals 3 locations for each cylinder with maximal taxel deflection when  a row of taxels aligns directly over the cylinder. Additionally, these PC vectors are close to each other irrespective of the class label, resulting in ambiguity, as is confirmed in our later analysis. 

Likewise, the BIOTACT whisker visualization (Fig.~1d, Figs.~\ref{PCA}d,h) shows that there is not a clear change in PC magnitude between cylinders, but simply a translation. This will also result in the cylinder curvature being more ambiguous than changes over location.

\subsection{Simple classification by nearest neighbour}
\begin{figure*}[t!]
	\vspace{0.6em}
	\centering
	\begin{subfigure}{0.2\textwidth}
		\centering
		\begin{overpic}[width=\textwidth]{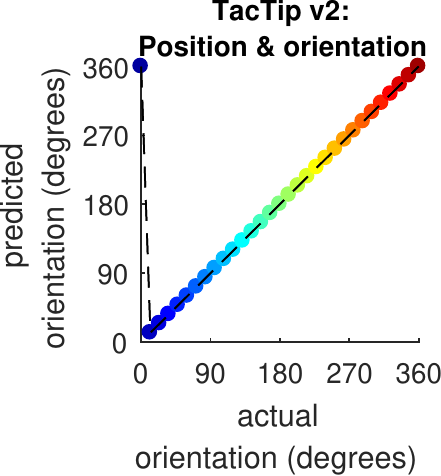}
			\put(5,95){\textbf{(a)}}
		\end{overpic}
		
	\end{subfigure}	\hfill
	\begin{subfigure}{0.2\textwidth}
		\centering
		\begin{overpic}[width=\textwidth]{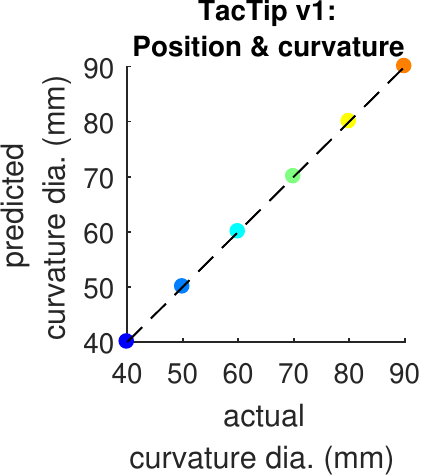}
			\put(5,95){\textbf{(b)}}
		\end{overpic}
	\end{subfigure}	\hfill
	\begin{subfigure}{0.2\textwidth}
		\centering
		\begin{overpic}[width=\textwidth]{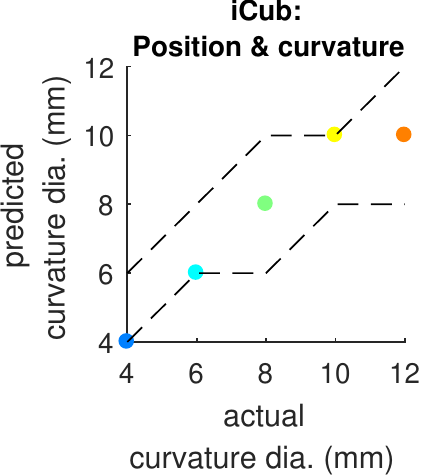}
			\put(5,95){\textbf{(c)}}
		\end{overpic}
	\end{subfigure}	\hfill
	\begin{subfigure}{0.2\textwidth}
		\centering
		\begin{overpic}[width=\textwidth]{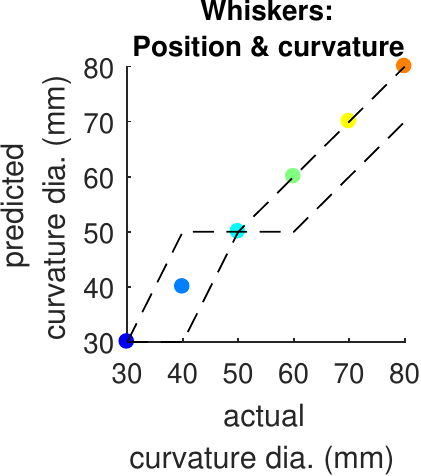}
			\put(5,95){\textbf{(d)}}
		\end{overpic}
	\end{subfigure}
	
		\vspace{0.5cm}
	\begin{subfigure}{0.2\textwidth}
		\centering
		\begin{overpic}[width=\textwidth]{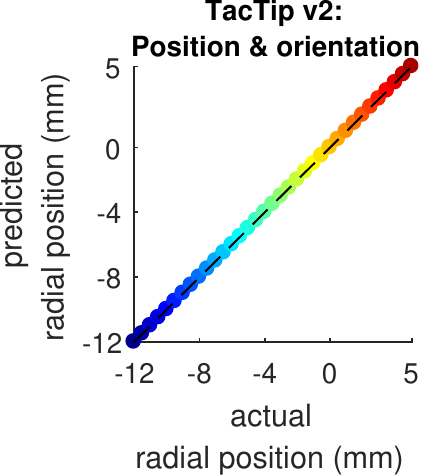}
			\put(5,97){\textbf{(e)}}
		\end{overpic}
	\end{subfigure}	\hfill
	\begin{subfigure}{0.2\textwidth}
		\centering
		\begin{overpic}[width=\textwidth]{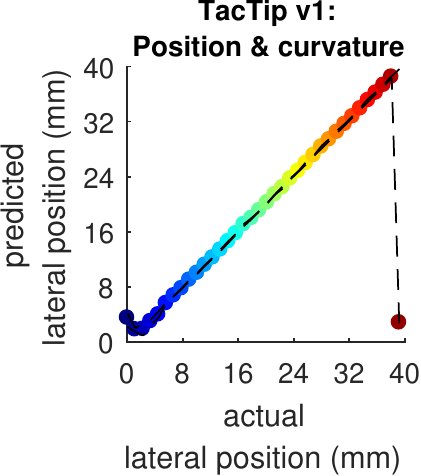}
			\put(5,97){\textbf{(f)}}
		\end{overpic}
	\end{subfigure}	\hfill
	\begin{subfigure}{0.2\textwidth}
		\centering
		\begin{overpic}[width=\textwidth]{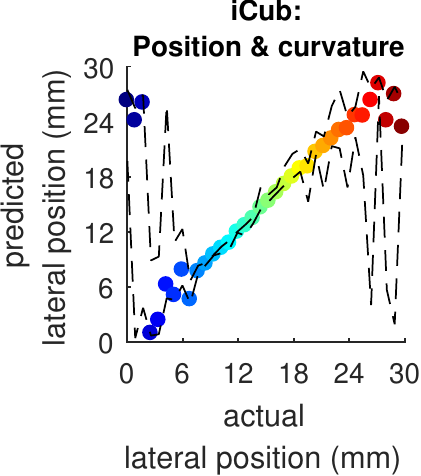}
			\put(5,97){\textbf{(g)}}
		\end{overpic}
	\end{subfigure}	\hfill
	\begin{subfigure}{0.21\textwidth}
		\centering
		\begin{overpic}[width=\textwidth]{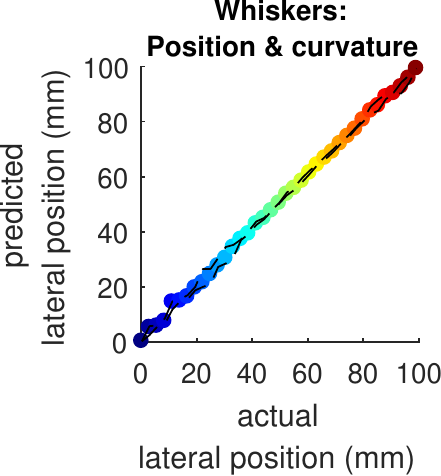}
			\put(5,97){\textbf{(h)}}
		\end{overpic}
	\end{subfigure}
	\caption{Nearest neighbour classification over PCs, showing classified vs actual classes. (a-d) Coloured markers show the median classified `what' identity class (orientation (a) or cylinder diameter, b-d) and dashed lines the 25\textsuperscript{th} and 75\textsuperscript{th} percentile classes. (e-h) Same quantities for the classified `where' location classes (radial position (e) and lateral position (f-h)). Closeness to the line of unit gradient indicates overall performance of the NN classifier. The colour bar from Fig.~\ref{PCA} is used.}
	\label{predict-actual1-what}
\end{figure*}

The structure of the data described above permits to use a very simple classifier (nearest neighbour; Sec. \ref{sec:methodsKnn}) to accurately perceive the `where' location and `what' stimulus identity. The classifier is verified by considering the inference results obtained for a separate collected test set. The total number of PCA dimensions $N_{\text{reduced}}$ kept for each set are: 5 for the TacTip v2 set, 5 for the TacTip v1 set, 4 for the iCub set and 6 for the BIOTACT Whiskers set. 

It can be immediately seen from Fig.~\ref{predict-actual1-what} that the classification results for the TacTip~v2 and TacTip~v1 have the least errors of all sensors considered. (One should note that the experiments are different, so this does not imply the sensors are superior.) This can be quantified by the gradient of the linear regressor of the predicted vs actual class results (assuming 0 intercept): which are 1.00, 0.99, 0.89, 0.96 for the `what' identity classification (Figs.~\ref{predict-actual1-what}a-d) and 1.00, 0.94, 0.91 and 1.00 for the `where' location classes (Figs.~\ref{predict-actual1-what}e-h). 

Considering the orientation graph, the errors for the TacTip~v2 `what' orientation classes (Fig.~\ref{predict-actual1-what}a) at angle 0\degree{} are due to 360\degree{} stimulus being equivalent to the 0\degree{} one, thereby justifying the output of the classifier.
The results for the iCub fingertip dataset (Fig.~\ref{predict-actual1-what}c) show that it is more difficult to distinguish between different cylinders (especially between the two largest cylinders with 10\,mm and 12\,mm diameter) than for cylinders in the TacTip~v1 and whiskers experiments. This is due to larger increments (10\,mm) for the TacTip~v1 and whiskers experiments.  

The `where' location class results show that the TacTip~v1 and iCub fingertip experiments have qualitatively similar behaviour (Figs.~\ref{predict-actual1-what}f,g), with both misclassifying data on the initial and final regions of contact with the cylinders where the sensor either had no or very light contact with the stimulus. The performance of the iCub fingertip experiment is worse because a large proportion of the considered range was not in contact with the cylinder. This misclassification is depicted by the vectors converging on (Figs.~\ref{PCA}f,g). Conversely, the whisker experiment (Fig.~\ref{predict-actual1-what}h) has less misclassification because the initial and final regions remain in contact with the cylinders and are distinct as seen in (Fig.~\ref{PCA}h). 

Finally, we compare the performance of the $k$-NN classifier with that of the probabilistic classifier (Sec.~\ref{sec:probClassifier}). Comparing the `what' perception root-mean-square errors obtained from the probabilistic classifier and the $k$-NN, one can conclude that both have comparable performance (\autoref{tbl:fixtable}). This shows that due to the structure present in the data as depicted by the visualisation graphs (Fig.~\ref{PCA}), even a simple classifier such as nearest neighbour can give satisfactory results.
 \begin{figure*}[t!]
 \vspace{0.5em}
	\centering
	
	\begin{subfigure}{0.21\textwidth}
		\centering
		\begin{overpic}[width=\textwidth]{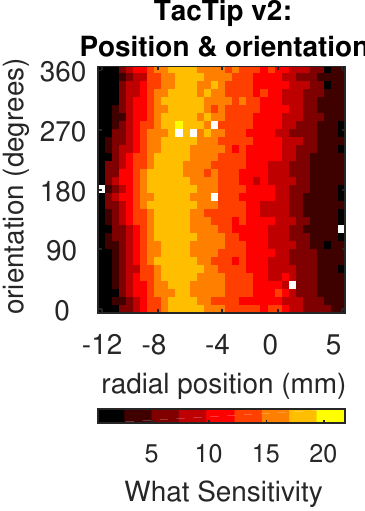}
			\put(5,91){\textbf{(a)}}
		\end{overpic}
		
	\end{subfigure}	\hfill
	\begin{subfigure}{0.2\textwidth}
		\centering
		\begin{overpic}[width=\textwidth]{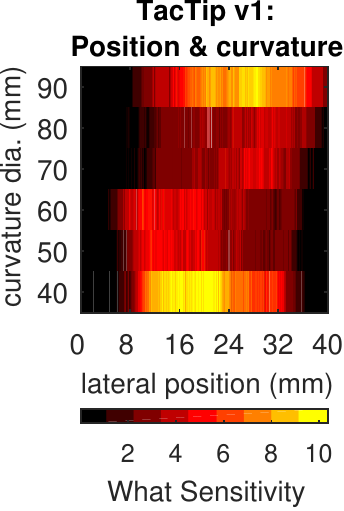}
			\put(4,90){\textbf{(b)}}
		\end{overpic}
		
	\end{subfigure}	\hfill
	\begin{subfigure}{0.2\textwidth}
		\centering
		\begin{overpic}[width=\textwidth]{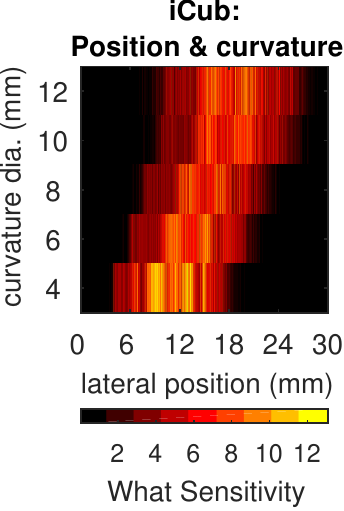}
			\put(4,90){\textbf{(c)}}
		\end{overpic}
		
	\end{subfigure}	\hfill
	\begin{subfigure}{0.2\textwidth}
		\centering
		\begin{overpic}[width=\textwidth]{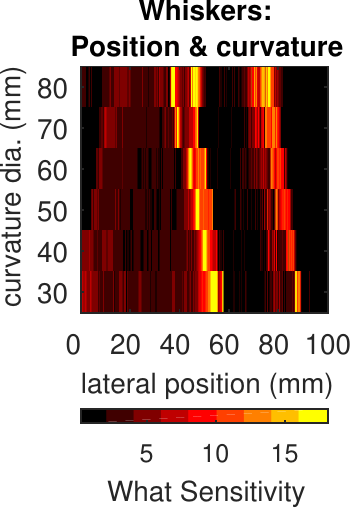}
			\put(4,90){\textbf{(d)}}
		\end{overpic}
		
	\end{subfigure}
	\caption{Sensitivity maps over the `what' identity and `where' location classes, derived from the tactile PCs. Only the normalised sensitivity (sensitivity as per \eqref{eq:sens} multiplied by the 'what' class increment shown in \autoref{tbl:Experiment data}) with respect to the `what' class is considered, derived from Figs.~\ref{PCA}a-d.}
	\label{heatmap}
		\vspace{0.35cm}
\end{figure*}
\begin{figure*}[t!]
	\centering
	\begin{subfigure}{0.22\textwidth}
		\centering
		\begin{overpic}[width=\textwidth]{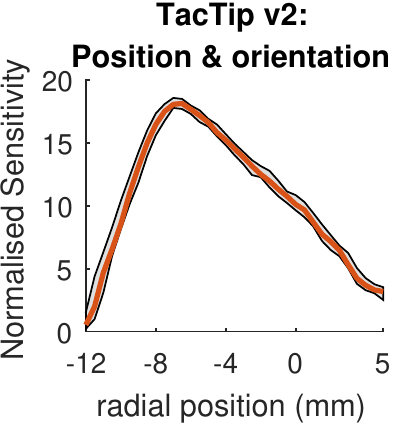}
			\put(4,92){\textbf{(a)}}
		\end{overpic}
		
	\end{subfigure}	\hfill
	\begin{subfigure}{0.23\textwidth}
		\centering
		\begin{overpic}[width=\textwidth]{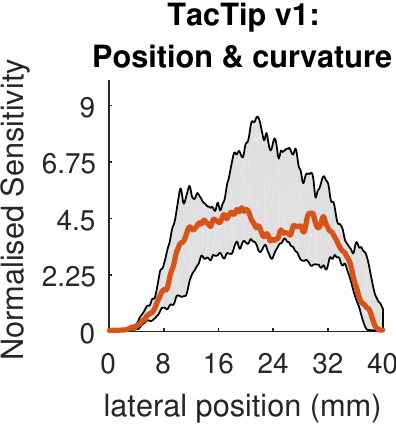}
			\put(4,92){\textbf{(b)}}
		\end{overpic}
		
	\end{subfigure}	\hfill
	\begin{subfigure}{0.23\textwidth}
		\centering
		\begin{overpic}[width=\textwidth]{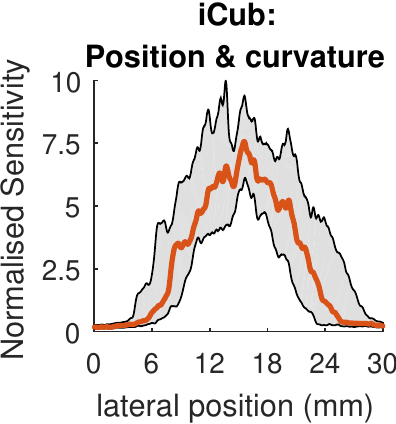}
			\put(4,92){\textbf{(c)}}
		\end{overpic}
	\end{subfigure}	\hfill
	\begin{subfigure}{0.23\textwidth}
		\centering
		\begin{overpic}[width=\textwidth]{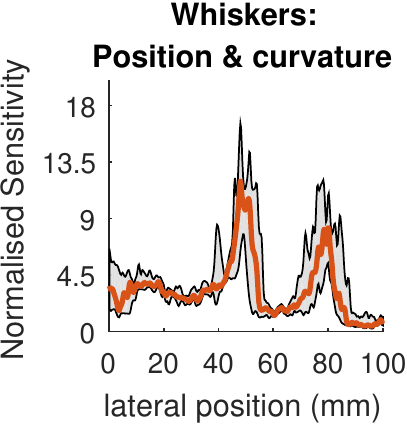}
			\put(4,92){\textbf{(d)}}
		\end{overpic}
	\end{subfigure}\hfill
	\caption{The median (orange) and 25\textsuperscript{th} and 75\textsuperscript{th} (black line) percentiles of the normalised sensitivity values at each `where' location class in Fig.~\ref{heatmap}. The plots are filtered to ease depiction and thus estimating the location of peak sensitivity.}
	\label{sensitivity1}
	\vspace{0.35cm}
\end{figure*}
\begin{figure*}[t!]
	\centering
	
	\begin{subfigure}{0.2\textwidth}
		\centering
		\begin{overpic}[width=\textwidth]{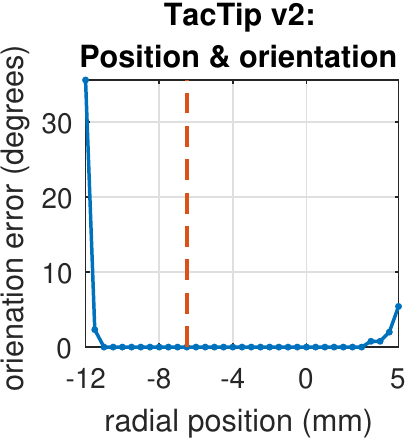}
			\put(4,92){\textbf{(a)}}
		\end{overpic}
	\end{subfigure}	\hfill
	\begin{subfigure}{0.21\textwidth}
		\centering
		\begin{overpic}[width=\textwidth]{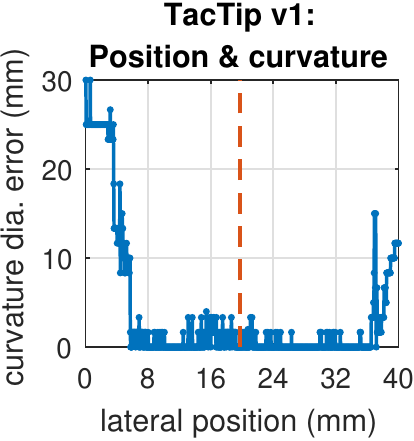}
			\put(4,92){\textbf{(b)}}
		\end{overpic}
	\end{subfigure}	\hfill
	\begin{subfigure}{0.21\textwidth}
		\centering
		\begin{overpic}[width=\textwidth]{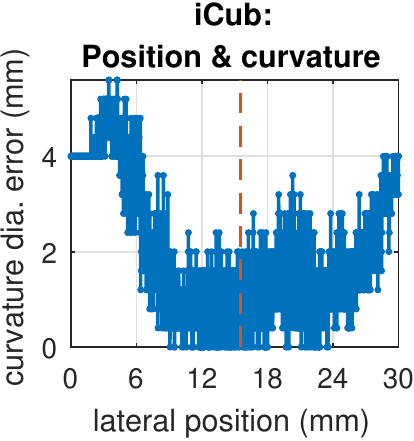}
			\put(4,92){\textbf{(c)}}
		\end{overpic}
	\end{subfigure}	\hfill
	\begin{subfigure}{0.2\textwidth}
		\centering
		\begin{overpic}[width=\textwidth]{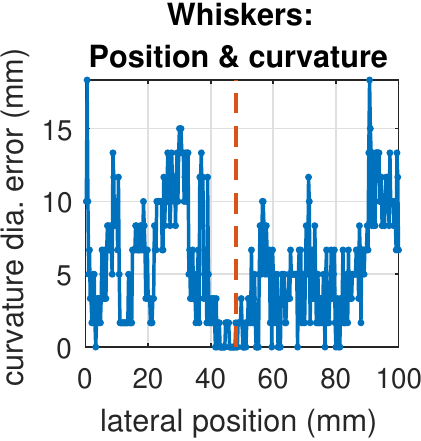}
			\put(4,92){\textbf{(d)}}
		\end{overpic}
	\end{subfigure}\hfill
	\caption{The `what' classes absolute error plots using the probabilistic classifier for (a) sensor orientation and (b-d) cylinder curvatures. The extremities of the location range shows a large error for all experiments. The iCub and whisker error plots (c-d) show a more complex error graph. The fixation locations presented in \autoref{tbl:fixtable} are shown as dashed lines.}
	\label{error}
\end{figure*}
\subsection{Sensitivity of Principal Components}

\begin{table}[t]
	\centering
	\caption{Summary of Results}
	\label{tbl:fixtable}
	\setlength{\tabcolsep}{0.4em}
	\begin{tabular}{ccccc}
		\hline
		    & TacTip v2 &TacTip v1 & iCub& Whiskers \\ \hline
		fixation `where' class no & 12 &493&1551& 192 \\
		fixation target value    &-6.5\,mm&19.72\,mm&15.51\,mm&48\,mm\\\hline
		`what' class error & 2.47\degree{} &5.55\,mm&2.56\,mm&7.95\,mm \\
		`what' class error ($k$-NN)    &0.98\degree{}& 2.67\,mm& 2.35\,mm&7.44\,mm\\\hline
	\end{tabular}
\end{table}

Another aspect of the structure of the PCs, is that it can be used to describe the distinguishability of the data to changes in `where' location and `what' identity class label. For example, very light contact results in a small sensitivity, since it is more difficult to distinguish between data that has small and thus very similar PC vectors. 

Here we depict these sensitivities of the PCs as heatmaps (Fig.~\ref{heatmap}), derived using the expression for sensitivity (\eqref{eq:sens}, Sec.\ref{sec:sensitivityMethod}) applied to the dimensionally reduced training set. We hypothesise that the dark regions (low sensitivity) reveal locations that result in larger object identity errors and the lighter regions (high sensitivity) show regions with better perception. In all sets, the regions at the extremities of the location ranges have larger errors, as expected since the sensors are in ambiguous locations such as free space. Curiously, for the whiskers, in addition to the extremities there is an internal region with low sensitivity (Fig.~\ref{heatmap}d), relating to ambiguous whisker data over those locations.

Additionally, the heatmaps are a good tool to visualise how the tactile contact data varies during an experiment. For example with TacTip~v2, the region with largest sensitivity is located between -8\,mm and -4\,mm (Fig.~\ref{heatmap}a), as also seen in the PC visualisation (Fig.~\ref{PCA}a). The heatmaps also show which `what' identity classes are the most distinct. For example, the smallest and largest cylinders are the most distinct for TacTip~v1 (Fig.~\ref{heatmap}b; see also Fig.~\ref{PCA}b). Additionally, the heatmaps clearly show when the magnitude of the taxel deflections becomes significant compared to no or light contact. For example, with the TacTip~v1 experiment, the sensor is in contact with the stimuli at approximately the same locations (8\,mm--32\,mm), whereas the iCub fingertip contacts at different starting points and ranges of locations. 

To give an overall indication of the sample population of sensitivity at each `where' location, we depict the median sensitivity and the 25\textsuperscript{th}/75\textsuperscript{th} percentiles of the `what' sensitivity values at each location (Fig.~\ref{sensitivity1}). We hypothesise that the peaks of these plots (equivalent to the light regions in Fig.~\ref{heatmap}) show the best locations to fixate the sensors for perceiving stimulus identity. The computed fixation points are presented in \autoref{tbl:fixtable}. A fixation point of $-6.5$\,mm off the stimulus edge is chosen for TacTip~v2. Curiously, the chosen fixation point is away from the edge or the middle of the stimulus. This occurred since at this location there is a good taxel mobility to perceive orientation (as discussed in Sec.~\ref{sec:resultVis}). On the other hand, for the cylinder experiments performed using the TacTip~v1, iCub fingertip and BIOTACT whiskers, a fixation location close to the centre of the cylinder is chosen (see \autoref{tbl:fixtable}). This is a reasonable choice since at that location the sensors have the largest taxel deflections.

Here we validate our hypothesis of the sensitivity showing the best regions for perception by considering inference results of a test set using a probabilistic classifier that takes the PCs as input (Sec.~\ref{sec:probClassifier}), rather than the raw sensor data~\cite{Lepora2016}. Crucially, the `what' perceptual errors variation with respect to the location `where' class (Fig.~\ref{error}) have the same behaviour as the sensitivity analysis (Fig.~\ref{heatmap} and Fig.~\ref{sensitivity1}). Therefore in regions with high (low) sensitivity the errors are low (high), validating our claim that useful information can be inferred about the experiment and tactile sensor simply by analysing a systematically-collected training set. 

\section{DISCUSSION}

This paper showed that the visualisation of various experiments with a range of different tactile sensors could be achieved by using a linear dimensionality reduction technique: PCA. The visualisations of the dimensionally-reduced training data were facilitated by collecting a systematic grid of stimulus locations and identities (`where' and `what'), revealing a surprising amount of structure and regularity that aid the interpretation of how a sensor interacts with a stimulus. Additionally, the uncovered structure in the data can help design simple machine learning methods that still obtain acceptable accuracies. For instance, we showed that the PC vector angle can be directly mapped onto actual sensor orientation (Fig.~\ref{PCA}a), showing the visualization can be used to design inference methods rather than merely being a tool to interpret results from other inference algorithms. 

The PC vectors for each tactile data segment $z$ lie on a manifold ${\cal M}_P$, the shape and nature of which provides further information about the sensor and experiment. A key measure was the sensitivity of each projected data segment, which quantifies how easy it is to distinguish between tactile data having different object identities at a specific location. The advantage of computing the sensitivity is twofold. Firstly, further understanding of the experiment is obtained from its visualisation and secondly a good point to locate the sensor for perception of the object can be computed from it. The fixation location produced from this simple method is then readily available to be included in a control policy for active perception or exploration~\cite{Lepora2016}.

An implication of this approach is that a simple inference procedure can then give the required perception data to control loops that require tactile feedback~\cite{Lepora2017,li2013}. Furthermore, the variation of tactile data within the low-dimensional manifold of PC vectors might also be used as a feedback signal to improve control. These implications fit with observations that a simple $k$-NN classifier within that manifold was sufficient to obtain direct, accurate estimates of the tactile sensor state, so that once the structure present in tactile data is uncovered, simple inference methods can be used for perception and control. 

To conclude, we note that in tactile robotics, action and perception are intrinsically tied together \cite{prescott2011}. In this study, these aspects of active perception are apparent in the useful information contained in the dimensionally-reduced manifold of tactile data. Because of the variety of tactile sensors and types of stimuli considered here, we expect that these observations will generalize to other sensors and other degrees-of-freedom, such as tactile sensors embedded in robot hands. As such, they offer the potential for novel control methods based on simple inference for in-hand manipulation and other tasks requiring tactile dexterity.


\end{document}